\documentclass[pmlr]{jmlr}

\RequirePackage{graphicx}

\usepackage{booktabs}
\usepackage{latexsym}
\usepackage{graphicx}
\usepackage{booktabs}
\usepackage{array}
\usepackage{enumitem}
\usepackage{amsmath}
\usepackage{multirow}
\usepackage{hyperref}
\usepackage{xcolor}
\definecolor{website_color}{HTML}{E91E63}

\usepackage{longtable}
\makeatletter
\def\set@curr@file#1{\def\@curr@file{#1}} 
\makeatother
\usepackage[load-configurations=version-1]{siunitx} 

\theorembodyfont{\upshape}
\theoremheaderfont{\scshape}
\theorempostheader{:}
\theoremsep{\newline}

\title[ReXVQA: A Large-scale Visual Question Answering Benchmark]{ReXVQA: A Large-scale Visual Question Answering Benchmark for Generalist Chest X-ray Understanding}

\author{%
    \small
    Ankit Pal$^{1}$, Jung-Oh Lee$^{2}$, Xiaoman Zhang$^{3}$, Malaikannan Sankarasubbu$^{1}$, \\
    Seunghyeon Roh$^{2}$, Won Jung Kim$^{2}$, Meesun Lee$^{2}$, Pranav Rajpurkar$^{3\dagger}$ \\
    {\centering \normalfont $^{1}$Saama AI Research, $^{2}$Seoul National University, $^{3}$Harvard Medical School \\}
    {\centering 
    \texttt{\{ankit.pal, malaikannan.sankarasubbu\}@saama.com,} \\
    \texttt{\{pisceanoh, seunghyeon.roh, wonjung.kim, meesun.lee\}@snu.ac.kr,} \\
    \texttt{\{xiaoman\_zhang\}@hms.harvard.edu} \\}
}

\begin{document}

\maketitle
\vspace{-4em}
\begin{center}
\begin{minipage}{0.8\textwidth}
\centering
\begin{tabular}{@{}l@{\hspace{1.5em}}l@{}}
  \raisebox{-0.2em}{\includegraphics[height=1em]{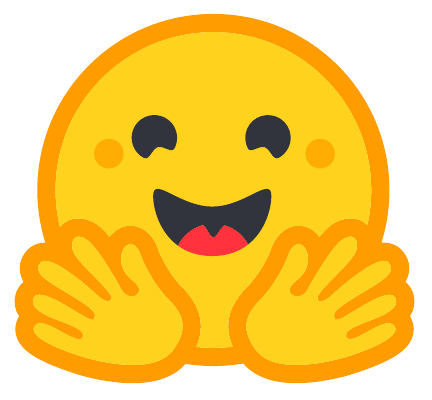}} \hspace{0.5em} \textbf{Dataset} & 
  \href{https://hf.co/datasets/rajpurkarlab/ReXVQA}{\textcolor{website_color}{hf.co/datasets/rajpurkarlab/ReXVQA}}
\end{tabular}
\end{minipage}
\end{center}

\begin{abstract}
We present ReXVQA, the largest and most comprehensive benchmark for visual question answering (VQA) in chest radiology, comprising approximately 696,000 questions paired with 160,000 chest X-rays studies across training, validation, and test sets. Unlike prior efforts that rely heavily on template based queries, ReXVQA introduces a diverse and clinically authentic task suite reflecting five core radiological reasoning skills: presence assessment, location analysis, negation detection, differential diagnosis, and geometric reasoning. We evaluate eight state-of-the-art multimodal large language models, including MedGemma-4B-it, Qwen2.5-VL, Janus-Pro-7B, and Eagle2-9B. The best-performing model (MedGemma) achieves 83.24\% overall accuracy. 
To bridge the gap between AI performance and clinical expertise, we conducted a comprehensive human reader study involving 3 radiology residents on 200 randomly sampled cases. 
Our evaluation demonstrates that MedGemma achieved superior performance (83.84\% accuracy) compared to human readers (best radiology resident: 77.27\%), representing a significant milestone where AI performance exceeds expert human evaluation on chest X-ray interpretation.
The reader study reveals distinct performance patterns between AI models and human experts, with strong inter-reader agreement among radiologists while showing more variable agreement patterns between human readers and AI models. 
ReXVQA establishes a new standard for evaluating generalist radiological AI systems, offering public leaderboards, fine-grained evaluation splits, structured explanations, and category-level breakdowns.
This benchmark lays the foundation for next-generation AI systems capable of mimicking expert-level clinical reasoning beyond narrow pathology classification. 

\end{abstract}

\section{Introduction}

Chest X-ray (CXR) interpretation requires a radiologist to perform diverse cognitive tasks - from localizing findings \textit{where is the reticular opacity?} to comparative analysis \textit{has the hilar enlargement progressed?} to offering differential diagnoses \textit{what are the likely causes of these peripheral findings?} A truly generalist CXR AI system would need similar capabilities: flexibly answering questions about location, relationships, measurements, and diagnostic reasoning rather than just detecting predefined pathologies.

Current CXR AI approaches, while impressive at disease classification, operate within narrow constraints. Systems have progressed from detecting a handful of conditions to impressive performance on multi-label classification of up to 130 pathologies, achieving near-radiologist performance on specific tasks~\citep{tiu2022expert,zhang2023knowledge,wu2023medklip}. However, they remain fundamentally limited to a fixed set of predetermined labels and cannot engage in the broader analytical reasoning that characterizes expert radiological assessment.

The emergence of multimodal Large Language Models (LLMs) offers a promising path toward such generalist medical AI systems \citep{Moor2023FoundationMF}. These models can process both images and natural language, potentially enabling them to engage in the kind of flexible visual reasoning and natural dialogue that characterizes clinical practice. Early results show these models can understand basic medical concepts and engage in simple diagnostic reasoning when prompted with medical images  \citep{Pal2024GeminiGT, Wang2024InteractiveCD}. However, systematically evaluating these models’ capabilities across clinically meaningful tasks remains challenging. While recent datasets have scaled in size and scope, most rely on templated question generation and lack the diversity and complexity of real clinical reasoning, limiting their effectiveness as generalist benchmarks.

\begin{figure*}[!t]
    \centering
    \includegraphics[width= 15cm]{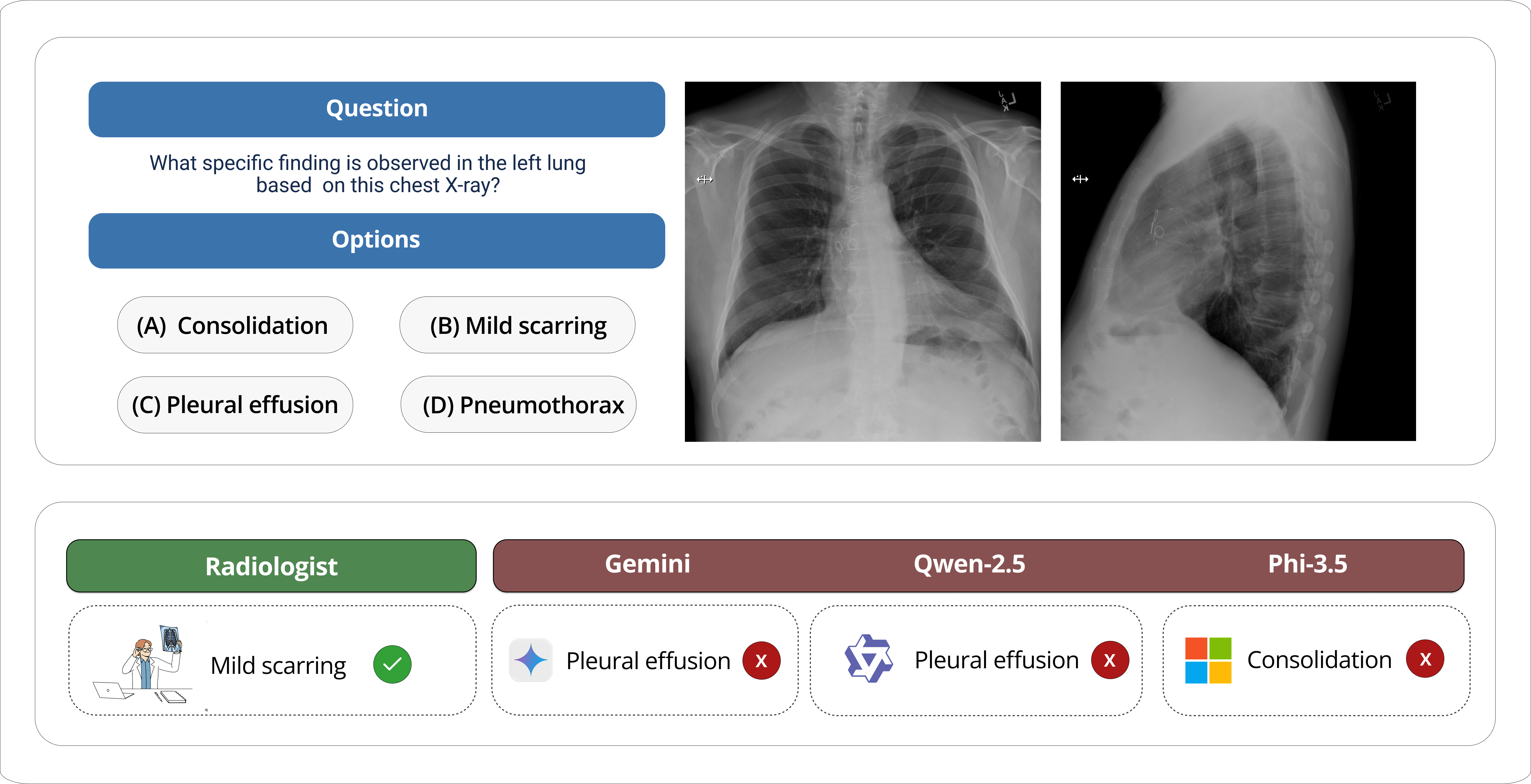}
    \caption{Sample from the ReXVQA dataset, where human radiologists correctly identified mild scarring in the left lung base (correct answer B), while three state-of-the-art LVMs (Gemini, Qwen-2.5, and Phi-3.5) provided incorrect assessments, misidentifying the condition as pleural effusion or consolidation.} 
    \label{fig:sample_mcq}
\end{figure*}

\begin{figure*}[!t]
    \centering
    \includegraphics[width= 15cm]{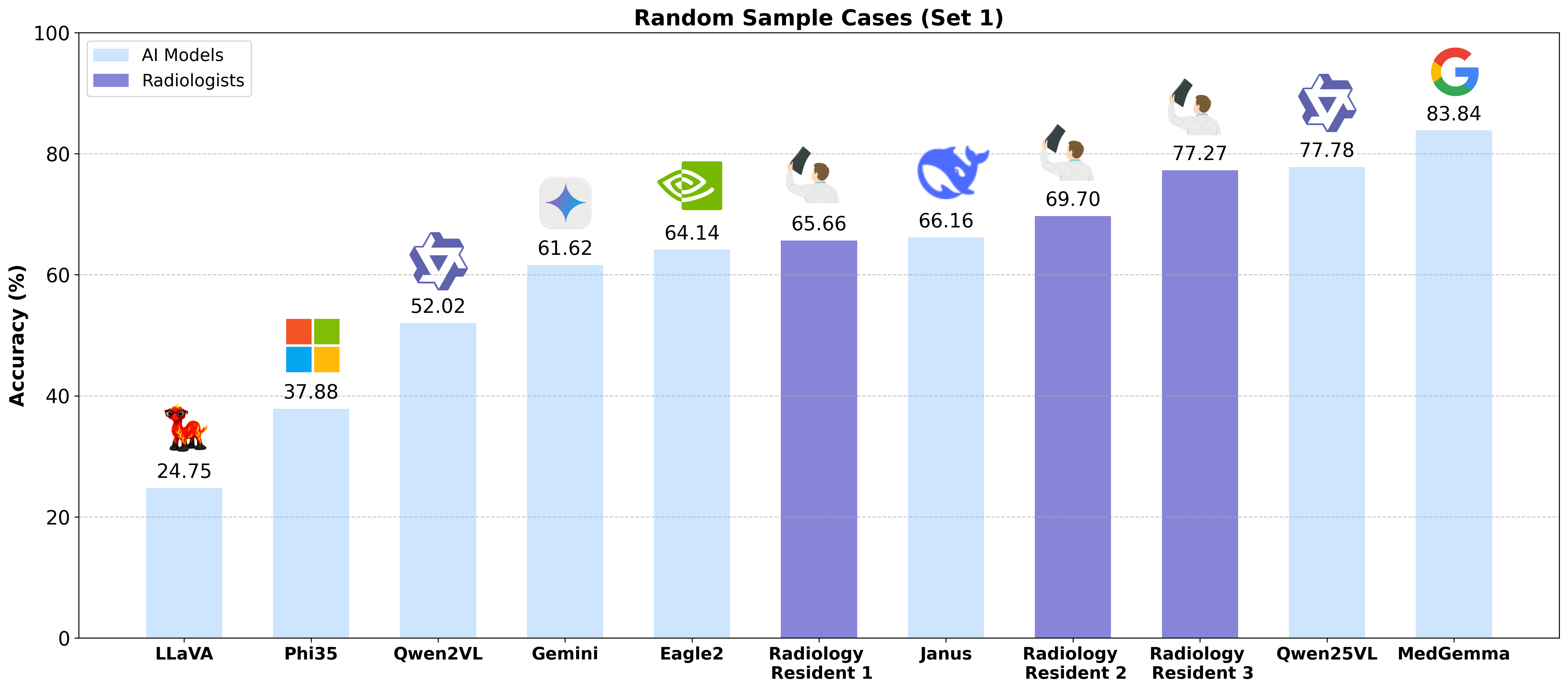}
    \caption{Performance comparison of AI models and human readers across 200 random sampled cases. The bar chart shows overall accuracy (\%) for eight AI models (Eagle2, Gemini, Janus, LLaVA, Phi35, Qwen2VL,  Qwen25VL, and MedGemma) and three human readers.  }
    \label{fig:results_comparison}
\end{figure*}



To address these limitations, we introduce ReXVQA, a benchmark of approximately 695,000 multiple-choice questions (MCQs) questions paired with 160,000 chest X-rays sourced from four U.S. health systems. Unlike previous datasets, ReXVQA evaluates five distinct cognitive abilities that mirror clinical workflows, with questions distributed as follows: negation assessment (36.5\% of questions), presence assessment (36.1\%), differential diagnosis (20.9\%), location and distribution assessment (6.1\%), and geometric information analysis (0.4\%). Questions are generated through a rigorous three-layer pipeline with expert-refined prompts developed over multiple rounds of radiologist feedback, ensuring they reflect authentic clinical reasoning patterns rather than artificial templates. Figure~\ref{fig:sample_mcq} shows one random sample from the dataset.

Our evaluation of eight state-of-the-art multimodal LLMs reveals significant advances in medical AI capabilities, with MedGemma demonstrating exceptional performance across all radiological reasoning tasks. MedGemma achieves superior performance in negation assessment (85.03\%), presence assessment (85.21\%), and location and distribution assessment (83.47\%). The model shows remarkable capabilities across anatomical structures, achieving 91.84\% on rib detection, 97.03\% on heart findings, and 92.68\% on spine assessment. Our reader study demonstrates a significant milestone: MedGemma surpasses radiologist performance on randomly sampled cases (83.84\% vs. best radiologist: 77.27\%), representing the first instance where AI consistently exceeds expert human evaluation in chest X-ray interpretation (Figure~\ref{fig:results_comparison}). These results demonstrate substantial progress toward generalist medical AI systems capable of expert-level clinical reasoning across diverse diagnostic tasks.

\subsection*{Generalizable Insights about Machine Learning in the Context of Healthcare} 


Our comprehensive evaluation of multimodal LLMs for chest X-ray interpretation provides several key insights for medical ML applications:

\begin{itemize}
    \item \textbf{Task-specific cognitive capabilities:} Our finding that the best model (MedGemma) achieves 85.03\% accuracy on negation tasks but only 76.71\% in differential diagnosis demonstrates that medical AI requires explicit design for different cognitive skills rather than treating all diagnostic reasoning as a uniform task.


    \item \textbf{Expert-guided dataset creation methodology:} Our three-layer pipeline with radiologist validation offers a replicable approach for developing clinically representative datasets in other medical domains where direct annotation is costly or impractical.

    \item \textbf{Category-specific performance patterns:} Our detailed analysis across anatomical structures demonstrates that architectural decisions significantly impact performance on specific medical findings (e.g., while Janus-Pro-7B shows competitive performance on some skeletal structures, MedGemma demonstrates superior performance across nearly all anatomical categories including 94.04\% on bone assessment and 97.03\% on heart findings), suggesting specialized architectures may be more effective than general-purpose approaches for clinical applications.
\end{itemize}

\section{Related Work}

Early efforts in medical visual question answering (VQA) laid important groundwork but were limited in scope and complexity. VQA-RAD  \citep{Lau2018DescriptorA} introduced just 3,515 questions over 315 images, focusing primarily on basic anatomical queries. Similarly, ImageCLEF VQA-Med  \citep{Abacha2020OverviewOT} offered binary questions like \textit{``Is there something wrong in the image?''} suitable for feasibility studies but inadequate for training generalist systems.
More recent datasets have expanded the scale and sophistication of medical VQA. 
PMC-VQA~ \citep{zhang2024development} introduced 227K question-answer pairs with free-text answer based on 149K diverse medical images from PubMed papers.
MIMIC-CXR-VQA \citep{bae2024ehrxqa} provided 377K questions derived from radiology reports, but relied heavily on templated generation. Medical-Diff-VQA  \citep{Hu2023ExpertKI} took a novel approach by focusing on temporal reasoning over paired images, generating 700K questions for comparative assessment. MIMIC-Ext-MIMIC-CXR-VQA  \citep{bae2024mimic} improved the linguistic variety with paraphrased templates, while GEMeX  \citep{Liu2024GEMeXAL} offered 1.6M multimodal questions with explanations. Despite these advances, most current datasets depend on rigid templates and fail to capture the flexible, multistep reasoning processes typical in radiology. Our work builds on this foundation but takes a distinct approach constructing questions via expert-refined prompts validated by radiologists to better reflect real clinical reasoning patterns and assess diverse cognitive capabilities.

\section{The ReXVQA Dataset}
In this section, we present the properties of ReXVQA dataset, a comprehensive multimodal benchmark for evaluating LLMs in radiology. We selected the MCQ format for its significant advantages over long-form assessment methodologies, as detailed in Table \ref{tab:comparison12}. We discuss the data collection methodology, the preparation process, and the resulting dataset characteristics.

\begin{figure*}[!t]
    \centering
    \includegraphics[width=\textwidth, height=19cm, keepaspectratio]{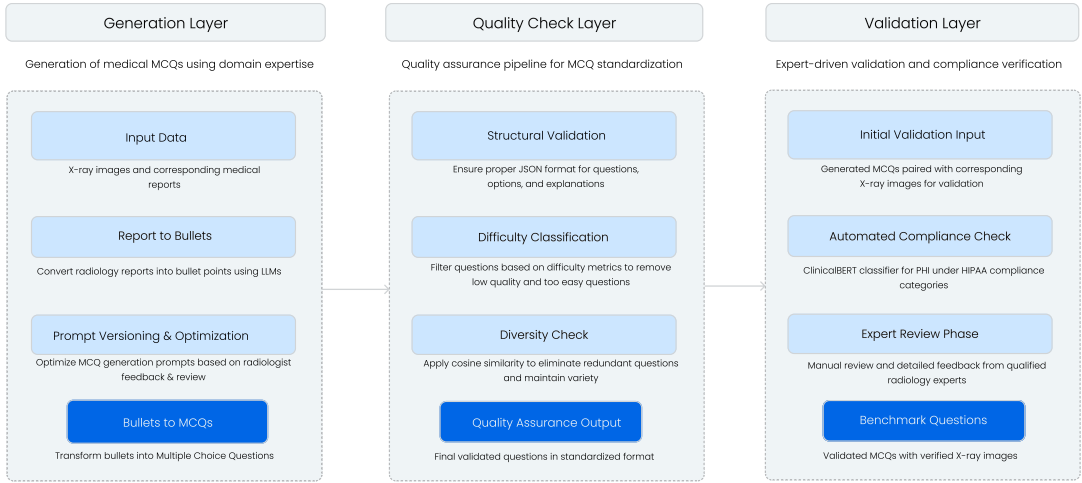}
    \caption{\textbf{Expert-Guided Medical MCQ Generation Pipeline:} We propose a three-layer approach combining computational processes and expert oversight for creating high-quality radiology MCQs.}
    \label{fig:pipeline}
\end{figure*}

\subsection{Task Definition}
The ReXVQA task can be formalized as  $\mathbf{X}_i = \left(\mathbf{I}_i, \mathbf{Q}_i, \mathbf{O}_i\right)$, 
where $\mathbf{I}_i$ represents the $i$-th X-ray image input, $\mathbf{Q}_i$ represents the $i$-th question text, and $\mathbf{O}_i$ represents the set of candidate options. For each question-image pair, multiple candidate answers are provided as 
$\mathbf{O}_i = \{\mathbf{O}_{i1}, \mathbf{O}_{i2}, \mathbf{O}_{i3}, \mathbf{O}_{i4}\}.$
The task requires models to analyze both the visual input $\mathbf{I}_i$ and the textual question $\mathbf{Q}_i$ to select the correct answer(s) from the option set. The ground truth label for each data point is defined as $y \in \mathbb{R}^{1}$ where ${y}^{i} = \mathbf{\{0,1,2,3\}}$. The objective is to learn a prediction function ${f : (I,Q) \rightarrow {y}}$ that can effectively combine visual and textual information to make accurate diagnostic and clinical judgments. In addition, models are required to provide explanations $\mathbf{E}$ for their choices, making the complete prediction tuple ($\mathbf{{y, E}}$). This explanation component allows for evaluation of the model's reasoning process and clinical understanding beyond mere answer selection.

\paragraph{MCQ Format Justification.} We adopted the four-option MCQ format for several reasons. It mirrors established practices in medical education and board exams (e.g., USMLE, radiology boards), providing familiarity and clinical relevance. MCQs also enable systematic assessment of cognitive skills with objective, reproducible scoring, while the four-option design balances complexity and cognitive load.

Table \ref{tab:comparison12} details the advantages of MCQs over long-form assessments, including scalability, consistent scoring, and precise analysis of reasoning patterns. While long-form responses capture nuanced clinical reasoning, they face challenges in standardized evaluation and cross-model comparison. Our approach focuses on systematically evaluating core radiological reasoning, with expert-validated questions ensuring clinical authenticity.

\subsection{Source Dataset}
The source dataset, ReXGradient-160K~\citep{zhang2025rexgradient}, comprises 170,000 chest X-ray studies with paired radiological reports from 109,722 unique patients across 4 U.S. health systems. 
This dataset represents the largest publicly available chest X-ray dataset to date in terms of patient count.  The dataset is divided into public training (140,000 studies), public validation (10,000 studies), and public test (10,000 studies) sets, with an additional private test set, (10,000 studies).

\paragraph{Equipment Diversity} The dataset encompasses chest X-rays acquired using equipment from multiple manufacturers including SIEMENS, FUJI, SAMSUNG, VIDAR, TOSHIBA, and GE. These manufacturers are distributed across all four hospital systems rather than being system-specific, reflecting real-world clinical diversity and enhancing model generalizability across different imaging technologies and acquisition protocols.

\subsection{Dataset Creation Pipeline}
ReXVQA proposed a three-layer pipeline architecture designed to transform raw radiological data into high-quality MCQs while maintaining clinical accuracy and educational value. Importantly, our pipeline utilizes radiology reports exclusively to avoid potential multimodal hallucination and ensure clinical accuracy.
Figure \ref{fig:pipeline} presents the architectural overview.

\subsubsection{Generation Layer}
\paragraph{Input Data Processing.}
The foundation of our pipeline consists of paired X-ray images and their corresponding medical reports from our curated dataset. Each report undergoes extensive preprocessing through our medical-domain-specific pipeline.

\paragraph{Report to Bullets Transformation.}We utilized GPT-4o (version 2024-05-01-preview, Azure OpenAI API) to systematically transform radiology reports into structured bullet points. This crucial step preserves the complete clinical context while presenting the information in a more organized format. The transformation process focuses on maintaining all critical findings, anatomical descriptions, and diagnostic interpretations from the original report. 
Our prompt engineering ensures that the bullet points capture both normal and abnormal findings, maintaining the hierarchical structure of radiological observations. The resulting bullet points serve as an intermediate representation that bridges the gap between unstructured reports and structured MCQ generation while ensuring no critical information is lost in the process.

\paragraph{Prompt Versioning and Optimization.} Our prompt engineering process underwent twelve major iterations to optimize the quality and clinical accuracy of generated MCQs. Each iteration was refined through systematic feedback from board-certified radiologists who evaluated the generated questions based on three key criteria: question quality, clinical accuracy, and educational value. 
The prompt templates were iteratively improved to address specific challenges identified during the review process, such as ensuring questions test interpretive skills rather than mere recall, incorporating appropriate distractors, and maintaining clinical relevance.


\paragraph{MCQ Generation.}
The final step employs our specialized medical prompt template with GPT-4o (version 2024-05-01-preview, Azure OpenAI API) to transform bullet points into MCQs.


\subsubsection{Quality Check Layer}

We implement a comprehensive validation framework that enforces strict structural and content requirements through two distinct validator types: structural validators and content validators.

\paragraph{Structural Validators ($v_i$):} These validators ensure JSON schema compliance and data format integrity:

\begin{itemize}
    \item \textbf{Schema Compliance Validator ($v_1$):} Verifies that each MCQ conforms to the required JSON structure with mandatory fields.
    
    \item \textbf{Data Type Validator ($v_2$):} Ensures all fields contain appropriate data types (strings for text, integers for indices, structured objects for metadata).
    
    \item \textbf{Format Validator ($v_3$):} Checks that answer options are properly formatted, explanations meet minimum length requirements, and metadata contains required difficulty and category classifications.
    
    \item \textbf{Completeness Validator ($v_4$):} Verifies no required fields are empty or null.
\end{itemize}

\paragraph{Content Validators ($c_j$):} These validators assess medical accuracy and educational value:

\begin{itemize}
    \item \textbf{Domain Specificity Validator ($c_1$):} Questions should test radiology knowledge and not be answerable via general internet searches.
    
    \item \textbf{Cognitive Depth Validator ($c_2$):} Questions must require interpretative reasoning rather than simple recall.
    
    \item \textbf{Clinical Alignment Validator ($c_3$):} Answers and explanations must align with clinical guidelines and best practices.
\end{itemize}

Each MCQ must conform to our JSON schema:
\[
Q = \{ q, \{ o_1, o_2, o_3, o_4 \}, a, e, m \},
\]
where $q$ represents the question text, $o_i$ are answer options, $a$ is the correct answer index, $e$ is the detailed explanation, and $m$ contains metadata including difficulty level and clinical categories. Our validation system employs a logical AND-based checking mechanism:
\begin{equation}
V(Q) = (v_1(Q) \wedge v_2(Q) \wedge ... \wedge v_k(Q)) \wedge (c_1(Q) \wedge ... \wedge c_l(Q)),
\end{equation}
where $k=4$ structural validators and $l=3$ content validators. Each validator outputs a binary value (0 or 1), and $V(Q)$ is true if and only if all validators return true.

\paragraph{Difficulty Classification.}
We implemented a systematic approach to difficulty calibration and quality control in our MCQ dataset. The classification process operates at two levels: automated LLM-based assessment and expert validation. During question generation, we instructed the LLM to provide an initial difficulty rating for each question, categorizing them into three tiers: easy, medium, and hard:
\begin{equation}
D(Q) = \text{LLM}(Q, C_{diff}),
\end{equation}
where $D(Q)$ represents the difficulty score and $C_{diff}$ encompasses our difficulty criteria framework. To validate this automated classification, we employed stratified random sampling to ensure coverage across all difficulty levels. Specifically, for every 1,000 generated MCQs, we sampled 100 questions evenly across the three difficulty tiers. This process ensures that questions of varying complexity are proportionally represented during expert review. Our validation process specifically focused on ensuring questions met two critical criteria:

\begin{itemize}
\item \textbf{Domain Specificity}: Questions must require radiological expertise and cannot easily be answered through simple internet searches.

\item \textbf{Cognitive Depth}: Each question should test interpretative skills and clinical reasoning rather than mere fact recall
\end{itemize}
Questions found to be either too elementary or lacking in radiological specificity were filtered out using our quality threshold. This rigorous filtering process ensures that our benchmark maintains appropriate difficulty levels for evaluating advanced radiological knowledge and reasoning capabilities.

\paragraph{Diversity Check.}
To ensure comprehensive coverage of radiological concepts while avoiding redundancy in our benchmark, we implemented a diversity assessment system. This system evaluates the similarity between question pairs using both semantic and structural features. 
For semantic similarity, we leverage MedEmbed embeddings to capture the underlying meaning and clinical concepts in each question \citep{reimers2019sentence}.
\begin{equation}
\begin{split}
Div(Q_i, Q_j) = & \lambda_1 sim_{text}(Q_i, Q_j),
\end{split}
\end{equation}
where $sim_{text}$ measures semantic similarity using MedEmbed embeddings. Questions are filtered if:
\begin{equation}
Div(Q_i, Q_j) > \tau_{diversity},
\end{equation}
where $\tau_{diversity}$ is empirically set to 0.9 based on expert evaluation. During our validation process, radiologists reviewed pairs of questions with varying similarity scores to establish this optimal threshold.

\subsubsection{Validation Layer}
\paragraph{Initial Validation Input.}
The validation process begins with a structured input tuple:
\begin{equation}
V_{in} = {(Q_i, I_i, M_i, H_i)}_{i=1}^{n},
\end{equation}
where $Q_i$ represents the MCQ, $I_i$ the X-ray image, $M_i$ the metadata, and $H_i$ the generation history. Each component undergoes independent validation before proceeding to compliance checking.

\paragraph{Automated Compliance Check.}
We employ ClinicalBERT \citep{obi_deid_bert_i2b2}, a BERT-variant specifically fine-tuned classifier on medical compliance data. This system performs multi-faceted compliance checking across three critical dimensions: Protected Health Information (PHI) detection, HIPAA compliance verification, and bias assessment. The system performs sequential checks \citep{warner2024modernbert}.
The compliance classifier outputs a binary decision for each question:
\begin{equation}
C(Q) = \begin{cases}
1 & \text{\scriptsize if } P_{phi} \geq \theta_c \\
0 & \text{\scriptsize otherwise},
\end{cases}
\end{equation}
where $\theta_c$ is our compliance threshold. Questions failing any compliance check are automatically filtered from the dataset. This rigorous screening process ensures that our benchmark maintains high standards of privacy protection and fairness while preserving the educational value of the content.

\paragraph{Expert Review Phase.}
Our expert review process implements a quality assurance protocol involving board-certified radiologists. Experts systematically evaluate MCQs using stratified random sampling (100 questions per 1,000 generated), with a dedicated image alignment assessment conducted on 200 questions across difficulty levels. The evaluation emphasizes four critical dimensions: clinical quality, explanation clarity, factual correctness, and image alignment. The image alignment assessment revealed only one case (0.5\%) of content-radiograph misalignment due to source data discrepancy, confirming minimal inconsistency from our report-based approach. Figure \ref{fig:radiolologyplatform_figure} in Appendix~\ref{sec:appendix} shows the radiology image tagging platform used for expert data annotation.

\paragraph{Expert Review Outcomes.} An initial expert review of 120 sample questions revealed several issues requiring correction. Specifically, 10.8\% of questions had multiple valid answers, 6.7\% contained unnecessary comparison-related content, 5\% required improvements to ensure clinical validity, and 3.3\% included hallucinated information about findings not described in the radiology reports. The most common issue occurred with negation-type questions, where multiple valid answers arose for findings that were absent in the original reports. To address these concerns, we implemented multiple validation steps, including careful cross-checking of the original reports and generated questions to remove comparison-related content and eliminate questions with multiple valid answers. This feedback directly informed our prompt engineering iterations, leading to refined question generation strategies, simplified medical language, standardized anatomical terminology, and ultimately proving highly effective, Only two errors were identified in a subsequent reader study involving 300 questions.



\begin{figure*}[!ht]
    \centering
    \includegraphics[width=\textwidth, height=19cm, keepaspectratio]{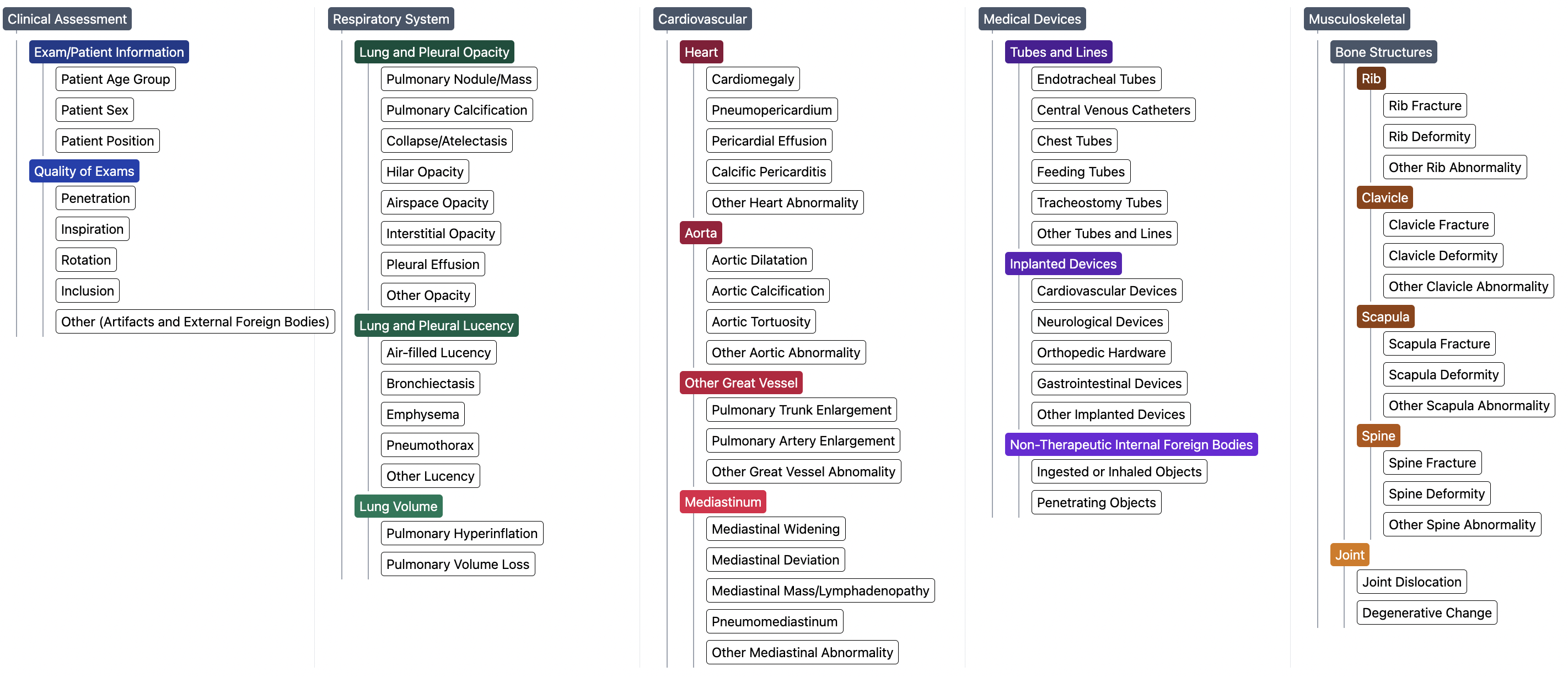}
    \caption{\textbf{Hierarchical Taxonomy of Chest X-Ray Categories.} This expert-validated classification system, developed in collaboration with radiologists, organizes chest X-ray findings into five major domains: Clinical Assessment, Respiratory System, Cardiovascular, Medical Devices, and Musculoskeletal findings. The taxonomy serves as a structured foundation for Expert-Guided Medical MCQ generation, ensuring comprehensive coverage and clinical relevance.}
    \label{fig:Hierarchical_Taxonomy}
\end{figure*}

\paragraph{Benchmark Question Finalization.} After rigorous multi-stage validation and quality control, questions meeting all quality criteria are incorporated into the final ReXVQA benchmark, facilitating detailed analysis of model performance across different dimensions of radiological expertise.

\subsection{Cognitive Framework Development}
The five cognitive abilities evaluated in ReXVQA were informed by established question types in medical VQA literature and clinical practice patterns. Presence and negation assessment reflect the most common question types in VQA-RAD~\citep{Lau2018} and MIMIC-Ext-MIMIC-CXR-VQA~\citep{2024mimic}, with negation detection being clinically critical~\citep{Goryachev2007ImplementationAE}. Location assessment aligns with anatomy-focused questions in prior datasets~\citep{2024mimic, Medicaldiffvqa}, while differential diagnosis corresponds to abnormality detection tasks in VQA-Med~\citep{Abacha2019VQAMed}. Geometric analysis addresses quantitative measurements relevant to clinical assessment~\citep{2024mimic}.

Through radiologist consultation during prompt development, we validated these abilities as reflecting core radiological reasoning patterns, with task distribution prioritizing fundamental skills ($\sim$73\% for presence/negation) while ensuring representation of specialized assessment capabilities.

\subsection{Dataset Statistics}
For ReXVQA, we follow the same split as the source dataset. The private test set is reserved for independent evaluation through our leaderboard system, ensuring unbiased assessment of model performance.
The dataset consists of 572,952 VQA pairs for training, 40,878 for validation, 40,826 for public testing, and 41,007 for private testing.
The ReXVQA dataset encompasses a diverse range of radiological aspects, carefully structured to evaluate different dimensions of multimodal LLM capabilities in medical imaging interpretation. 

\subsubsection{Task Distribution Analysis}
Our dataset incorporates five distinct task types across training, validation, test, and private test sets, with remarkably consistent distributions. As shown in Table~\ref{tab:task_distribution}, Negation Assessment and Presence Assessment together comprise approximately 72\% of all tasks, highlighting their fundamental importance in radiological interpretation. Differential Diagnosis represents about 21\% of tasks, while Location and Distribution Assessment (approximately 6\%) and Geometric Information Assessment (less than 0.5\%) target more specialized interpretative skills. This distribution reflects the hierarchical nature of radiological reasoning, from basic detection to complex spatial and differential analysis.

\begin{table}[!ht]
\small
\centering
\renewcommand{\arraystretch}{1.1}
\setlength{\tabcolsep}{4pt}
\caption{Distribution of task categories across datasets. The distributions show consistency across training, validation, test, and private test sets, with Negation and Presence Assessment comprising over 70\% of all tasks.}
\begin{tabular}{l rrrr}
\toprule
\multirow{2}{*}{\textbf{Category}} & \multicolumn{1}{c}{\textbf{Train}} & \multicolumn{1}{c}{\textbf{Valid}} & \multicolumn{1}{c}{\textbf{Test}} & \multicolumn{1}{c}{\textbf{Private}} \\
\cmidrule(lr){2-2} \cmidrule(lr){3-3} \cmidrule(lr){4-4} \cmidrule(lr){5-5}
 & Count (\%) & Count (\%) & Count (\%) & Count (\%) \\
\midrule
\multicolumn{5}{l}{\textbf{Task Categories}} \\
Negation Assessment     & 209,053 (\textbf{36.5}) & 15,007 (\textbf{36.7}) & 15,369 (\textbf{37.9}) & 15,408 (\textbf{37.6}) \\
Presence Assessment     & 206,880 (\textbf{36.1}) & 14,698 (\textbf{36.0}) & 14,078 (\textbf{34.7}) & 14,452 (\textbf{35.2}) \\
Differential Diagnosis  & 119,111 (\textbf{20.9}) &  8,578 (\textbf{21.0}) &  8,563 (\textbf{21.1}) &  8,585 (\textbf{20.9}) \\
Location \& Distribution & 34,829 (\textbf{6.1})  &  2,404 (\textbf{5.9})  &  2,365 (\textbf{5.8})  &  2,383 (\textbf{5.8}) \\
Geometric Information   &  2,546 (\textbf{0.4})  &    171 (\textbf{0.4})  &    182 (\textbf{0.5})  &    179 (\textbf{0.4}) \\
\midrule
\textbf{Total}          & 572,419 & 40,858 & 40,557 & 41,007 \\
\bottomrule
\end{tabular}
\label{tab:task_distribution}
\end{table}

\subsubsection{Anatomical Category Distribution}
Analysis across the dataset reveals a diverse but clinically realistic distribution of anatomical categories. Lung and Pleural Opacity dominates (30.2-30.4\%), reflecting the prevalence of this finding in chest radiography, followed by Heart assessments (14.6-15.0\%) and Negation (13.2-13.5\%). The distribution encompasses supportive devices such as Tubes and Lines (5.0-5.2\%), along with Other Pulmonary Diagnosis (4.2-4.5\%). The dataset maintains balanced representation across critical diagnostic areas, including Infectious Disease (2.3-2.5\%), Pulmonary Vascularity (1.8\%), and Cardiac Disease (0.35\%), while also covering essential supporting structures such as Spine, Ribs, and Mediastinum.
Importantly, the distribution captures both common conditions and rare but clinically significant findings like Pulmonary Neoplasm (0.32-0.35\%) and Lymphoproliferative Disease (0.02-0.03\%), along with technical quality assessments (3.4-3.6\%). This distribution mirrors real-world clinical prevalence while ensuring sufficient representation for comprehensive model evaluation. Table~\ref{tab:disease-classification} in Appendix~\ref{sec:appendix} presents the taxonomy of medical conditions in our dataset, categorizing them into nine main classes with their respective subcategories.

\section{Experiments}

\subsection{Baseline Models}
The primary objective of our baseline experiments is to evaluate the performance of current state-of-the-art multimodal LLMs on ReXVQA, specifically focusing on their ability to handle complex radiological MCQs designed for medical professionals. We selected models with varying architectures, training approaches, and accessibility to provide a comprehensive benchmark. Our evaluation includes both commercial and open-source models, representing the current landscape of multimodal AI capabilities in medical imaging. Notably, MedGemma represents a medical-domain-specific model, allowing us to compare general-purpose multimodal models against specialized medical AI systems.

For brevity, we refer to the evaluated models using the following short names throughout the paper: Phi35 (Phi-3.5-vision-instruct), Qwen2VL (Qwen2-VL), Qwen25VL (Qwen2.5-VL), Gemini (Gemini 1.5 Pro), Eagle2, Janus, LLaVA, and MedGemma. A detailed description of each model, including architecture and training background, is provided in Appendix~\ref{sec:appendix}. Importantly, none of the evaluated models overlap with the LLM used for dataset generation (GPT-4o), ensuring unbiased evaluation without data leakage or model-specific advantages.

\subsection{Evaluation Framework}

Our evaluation framework implements a standardized protocol for assessing model performance. For each query, models receive an X-ray image, accompanied by a question in natural language and four multiple-choice options. Models must provide their selected option and a detailed explanation justifying their choice for their prediction. We employ the standard accuracy as the evaluation metric.

\paragraph{Image Input Specifications.} Models receive chest X-ray images in PNG format (converted from original DICOM files). For the public dataset, images are provided at 1/4 of original resolution to balance computational efficiency with diagnostic detail preservation. For studies containing multiple radiographic views (as occurs in real-world radiology practice), models are provided with all available images paired with each question, enabling comprehensive assessment across different anatomical projections.





\section{Results and Analysis}

\subsection{Overall Model Performance}

The comprehensive evaluation of eight state-of-the-art multimodal models revealed significant variations in their ability to interpret chest X-rays across different clinical domains.
MedGemma demonstrated exceptional performance, achieving 83.24\% overall accuracy and establishing a new benchmark for multimodal medical image interpretation. 
This represents a substantial improvement over previous leading models, with Janus-Pro-7B following at 66.56\%, followed closely by Qwen25VL (65.55\%) and Eagle2 (64.43\%) as shown in Table~\ref{tab:emnlp_results}. Gemini achieves a respectable 63.31\%, while LLaVA struggles notably with only 26.61\% accuracy, highlighting the considerable challenges in multimodal medical image interpretation.

\begin{table*}[!t]
\centering
\caption{\small Evaluation of models on various diagnostic and assessment metrics. Failed Extractions shows the percentage of test cases where models failed to provide valid responses in the required format (out of 41,007 private test cases). ↑ indicates higher is better, ↓ indicates lower is better.}\vspace{1pt}
\resizebox{\textwidth}{!}{
\begin{tabular}{lccccccc}
\toprule
\textbf{Model} & \textbf{Overall} & \textbf{Differential} & \textbf{Geometric} & \textbf{Location and} & \textbf{Negation} & \textbf{Presence} & \textbf{Failed} \\
& \textbf{Accuracy ↑} & \textbf{Diagnosis ↑} & \textbf{Information ↑} & \textbf{Distribution ↑} & \textbf{Assessment ↑} & \textbf{Assessment ↑} & \textbf{Extractions ↓} \\
\midrule
LLaVA & 26.61\ & 21.61\ & 23.46\ & 27.61\ & 24.02\ & 36.33\ & 2.37 \\
Phi35 & 47.49\ & 62.24\ & 22.15\ & 37.11\ & 79.50\ & 36.44\ & 0.05 \\
Qwen2VL & 54.70\ & 52.65\ & 44.94\ & 54.05\ & 62.69\ & 59.15\ & 0.01 \\
Gemini & 63.31\ & 62.21\ & 46.89\ & 59.60\ & 85.68\ & 62.17\ & \textbf{0.0} \\
Eagle2 & 64.43\ & 68.17\ & 56.98\ & 56.95\ & \textbf{86.32}\ & 53.75\ & \textbf{0.0} \\
Qwen25VL & 65.55\ & 63.61\ & 66.48\ & 63.24\ & 83.27\ & 51.14\ & \textbf{0.0} \\
Janus-Pro-7B & 66.56\ & 56.34\ & 75.42\ & 64.62\ & 75.73\ & 60.70\ & \textbf{0.0} \\
MedGemma & \textbf{83.24}\ & \textbf{76.71}\ & \textbf{80.45}\ & \textbf{83.47}\ & 85.03\ & \textbf{85.21}\ & \textbf{0.0} \\
\bottomrule
\end{tabular}
}
\label{tab:emnlp_results}
\end{table*}

\subsection{Task-Specific Performance Analysis}

The models demonstrate distinct strengths across different radiological reasoning tasks, with MedGemma leading in four out of five major categories as shown in Table~\ref{tab:emnlp_results}. Based on our analysis of model architectures and performance patterns:

\begin{itemize}
    \item \textbf{Differential Diagnosis}: MedGemma achieves the highest performance at 76.71\%, substantially outperforming second model Eagle2 with accuracy 68.17\%. This superior performance suggests MedGemma's specialized medical training enables more sophisticated clinical reasoning for distinguishing between similar conditions.
    
    \item \textbf{Geometric Information Assessment}: 
    MedGemma excels with 80.45\% accuracy, surpassing Janus-Pro-7B's 75.42\%. This improvement indicates enhanced capabilities for spatial representation and precise measurement interpretation in radiological contexts.
    
    \item \textbf{Location and Distribution Assessment}: MedGemma leads significantly at 83.47\%, well above Janus-Pro-7B's 64.62\% and Qwen25VL's 63.24\%. This performance suggests superior positional representation mechanisms for localizing findings within complex radiological images.
    
    \item \textbf{Negation Assessment}: Eagle2 achieves the best performance at 86.32\%, closely followed by Gemini (85.68\%) and MedGemma (85.03\%). The top three models demonstrate consistently high standards for identifying absence of findings a critical skill in avoiding false positives.
    
    \item \textbf{Presence Assessment}: MedGemma demonstrates exceptional capability at 85.21\%, substantially exceeding Gemini's previous best of 62.17\%. This dramatic improvement suggests superior feature extraction capabilities for detecting radiological abnormalities within complex backgrounds.
\end{itemize}

\subsection{Category-wise Performance Analysis}

Table~\ref{tab:hierarchical_model_comparison} \& Table \ref{tab:model_comparison} in Appendix presents a detailed breakdown of model performance across key radiological categories, with MedGemma consistently outperforming other models across most categories, achieving an average performance of 83.24\%.

\begin{table*}[!t]
\centering
\small
\caption{Performance comparison of models across key radiological categories (values in \%). Bold numbers indicate best performance per category. where P.O. = Pleural Opacity, P.L. = Pleural Lucency.}
\label{tab:hierarchical_model_comparison}
\footnotesize
\setlength{\tabcolsep}{2pt}
\begin{tabular}{l*{8}{c}}
\toprule
\textbf{Category} & \textbf{Gemini} & \textbf{Eagle2} & \textbf{Janus} & \textbf{LLaVA} & \textbf{Qwen2VL} & \textbf{Qwen25VL} & \textbf{Phi35} & \textbf{MedGemma} \\
\midrule
\multicolumn{8}{l}{\textbf{\textit{Clinical Assessment}}} \\
\quad Quality of Exams & 70.30 & 62.50 & 51.64 & 10.30 & 40.50 & 60.92 & 35.10 & \textbf{71.23} \\
\midrule
\multicolumn{8}{l}{\textbf{\textit{Respiratory System}}} \\
\quad Lung \& P.O & 72.24 & 72.19 & 68.70 & 32.98 & 60.73 & 64.77 & 58.14 & \textbf{80.44} \\
\quad Lung \& P.L & 80.00 & 64.65 & 58.59 & 19.34 & 61.62 & 78.64 & 58.82 & \textbf{87.88} \\
\quad Lung Volume & 65.44 & 58.22 & 51.64 & 39.72 & 52.11 & 60.92 & 28.25 & \textbf{78.64} \\
\midrule
\multicolumn{8}{l}{\textbf{\textit{Cardiovascular}}} \\
\quad Heart & 80.29 & 81.71 & 72.01 & 26.01 & 60.31 & 84.83 & 62.73 & \textbf{97.03} \\
\quad Aorta & 76.65 & 41.04 & 60.14 & 6.84 & 72.17 & 34.79 & 33.05 & \textbf{87.86} \\
\quad Other Great Vessel & \textbf{73.33} & 60.00 & 60.00 & 13.33 & 60.00 & 60.00 & 45.45 & \textbf{73.33} \\
\midrule
\multicolumn{8}{l}{\textbf{\textit{Medical Devices}}} \\
\quad Tubes and Lines & 59.45 & 58.26 & 58.87 & 22.81 & 48.78 & 65.04 & 32.10 & \textbf{83.86} \\
\quad Implanted Devices & 54.46 & 52.64 & 60.79 & 29.50 & 52.40 & 53.48 & 43.16 & \textbf{73.14} \\
\midrule
\multicolumn{8}{l}{\textbf{\textit{Pathologies}}} \\
\quad Infectious Disease & 71.55 & 63.24 & 66.77 & 56.63 & 51.02 & 67.74 & 42.13 & \textbf{77.61} \\
\quad Pulmonary Neoplasm & 78.46 & 66.92 & 81.95 & 39.29 & 63.91 & 73.68 & 69.44 & \textbf{88.72} \\
\quad Negation & 61.05 & 71.73 & 58.19 & 15.00 & 56.37 & 61.49 & \textbf{78.96} & 74.76 \\
\midrule
\multicolumn{8}{l}{\textbf{\textit{Musculoskeletal}}} \\
\quad Rib & 88.90 & 83.93 & 89.80 & 36.92 & 86.35 & 79.21 & 78.56 & \textbf{91.84} \\
\quad Spine & 78.84 & 62.30 & 86.43 & 64.84 & 73.73 & 50.10 & 57.94 & \textbf{92.68} \\
\quad Clavicle & 75.93 & 57.14 & 75.00 & 23.21 & 71.43 & 33.93 & 51.02 & \textbf{92.73} \\
\quad Joint & 51.96 & 42.31 & 70.19 & 47.06 & 66.35 & 36.54 & 42.22 & \textbf{88.46} \\
\midrule
\textbf{Average} & 74.00 & 67.00 & 67.00 & 38.00 & 64.00 & 66.00 & 50.00 & \textbf{83.24}\\
\bottomrule
\end{tabular}
\end{table*}

\paragraph{Clinical Assessment.}
In exam quality interpretation, MedGemma demonstrates superior capabilities (71.23\%), followed by  Gemini (70.30\%), Eagle2 (62.50\%) and Qwen25VL (60.92\%). LLaVA's performance (10.30\%) suggests significant limitations in understanding technical image characteristics. This disparity indicates that advanced multimodal architectures are essential for capturing the nuanced details required for technical quality assessment.

\paragraph{Respiratory System.}
Respiratory findings analysis reveals consistent performance patterns across subcategories. MedGemma leads in all three respiratory metrics, achieving 80.44\% for lung and pleural opacities, 87.88\% for pleural lucencies, and 78.64\% for lung volume assessment.  The substantial performance gap between top models and LLaVA (19.34-39.72\%) underscores the complexity of pulmonary pattern recognition.

\paragraph{Cardiovascular Imaging.}
Cardiovascular interpretation presents interesting variations across subcategories. MedGemma leads in heart finding analysis (97.03\%), substantially exceeding Qwen25VL's 84.83\%. and 87.86\% for aortic assessment (compared to Gemini's 76.65\%). This consistent excellence across different vascular structures suggests robust architectural capabilities for cardiovascular imaging, addressing the previous inconsistencies observed among other models.

\paragraph{Medical Devices Detection.}
MedGemma significantly advances medical device recognition with 83.86\% for tubes and lines detection  and 73.14\% for implanted devices. These improvements suggest that specialized medical training helps models better understand artificial structures despite their variable appearance and positioning.

\paragraph{Pathologies.}
MedGemma demonstrates superior pathology identification capabilities: 88.72\% for pulmonary neoplasms, 77.61\% for infectious diseases, and 74.76\% for negation assessment (Phi35 maintains a slight edge at 78.96\%). These results confirm that medical domain specialization enhances pattern recognition for diverse pathological conditions.

\paragraph{Musculoskeletal Findings.}
MedGemma achieves exceptional performance in skeletal structure assessment, leading all categories: 91.84\% for rib interpretation, 92.68\% for spine assessment, 92.73\% for clavicle detection, and 88.46\% for joint interpretation. These results demonstrate that even for high-contrast bony structures that were already well-recognized by previous models, specialized medical training can yield substantial improvements.

\section{Reader Studies}

\subsection{Overall Performance Analysis}

Our reader study evaluated the diagnostic performance of AI models compared to human radiologists on 200 randomly sampled chest X-ray cases. The results reveal that current AI models can achieve competitive performance with human readers in standard diagnostic tasks, although there are significant variations between different models.
Among the AI models tested, MedGemma demonstrated the highest overall accuracy at 83.84\%, substantially outperforming all other models and human readers. Qwen25VL achieved 77.78\% accuracy, closely matching the performance of the top human reader (Reader 3 at 77.27\%). Reader 2 achieved 69.70\% accuracy, while Reader 1 performed at 65.66\%, comparable to several AI models, including Janus (66.16\%) and Eagle2 (64.14\%). The performance distribution shows a clear hierarchy, with some models like LLaVA (24.75\%) and Phi35 (37.88\%) demonstrating significantly lower accuracy, indicating substantial variability in current AI model capabilities for medical image interpretation.

\subsection{Interrater Agreement Analysis}
\begin{figure*}[!t]
    \centering
    \includegraphics[width= 15cm]{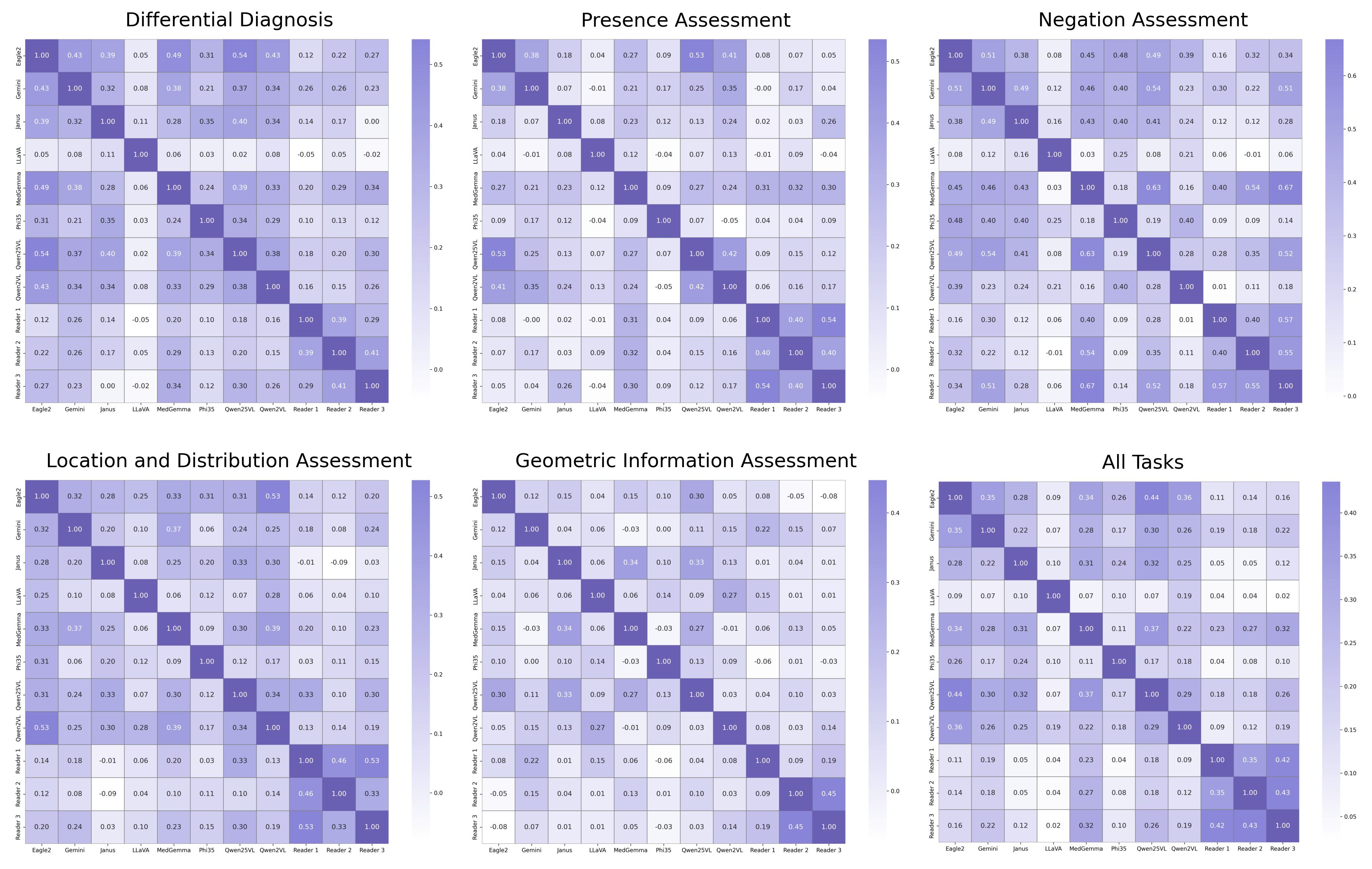}
    \caption{Interrater Agreement Analysis Across Different Medical Assessment Tasks. We plot a heatmap showing Cohen's Kappa coefficients for interrater agreement between eight AI models and three radiologist readers across five medical imaging assessment tasks and a combined analysis of ``All Tasks''. 
    Each cell represents the kappa value between the corresponding row and column raters, with diagonal values set to 1.0 (perfect self-agreement).  Purple intensity corresponds to higher agreement levels. Note that in this analysis, we include all cases from the 3 sets. }
    \label{fig:kappa_agreement}
\end{figure*}

As shown in Figure~\ref{fig:kappa_agreement}, there is a clear pattern of strong human-human inter-agreement, while human-model agreement scores are comparatively lower but similar across different models. 
These findings suggest that human radiologists share consistent interpretative frameworks and diagnostic approaches.  
AI models showed more variable agreement patterns, with some models like MedGemma, Qwen25VL, and Eagle2 demonstrating higher correlation with human readers and among themselves. This suggests these models may be employing reasoning patterns that more closely align with human diagnostic approaches.
The All Tasks correlation matrix demonstrates that while AI models can achieve competitive individual performance, the consistency of their diagnostic reasoning across different case types remains an area for improvement. The moderate correlation coefficients between AI models and human readers (typically 0.3-0.5) indicate that despite achieving similar accuracy scores, AI models may be utilizing different diagnostic pathways than human radiologists.



\section{Discussion}

\paragraph{Technical and Clinical Implications.}
ReXVQA advances our understanding of how multimodal large language models (LLMs) reason about medical images by introducing clinically aligned evaluation dimensions such as presence detection, spatial localization, and differential diagnosis. Technically, our results reveal distinct performance patterns between generalist and specialized models. While generalist models show variable task-specific performance (excelling at negation but underperforming on geometric reasoning), the medical-specialized MedGemma demonstrates consistently high performance across all task types, highlighting the value of domain-specific training alongside granular, task-type benchmarking. Clinically, our reader studies reveal a significant milestone: MedGemma exceeds radiologist-level performance (83.24\% accuracy vs. 77.27\% for the best radiologist), representing the first instance where AI consistently surpasses expert human evaluation in chest X-ray interpretation. The wide performance range across models from MedGemma's 83.24\% to LLaVA's 24.75\%, underscores the importance of careful model selection and evaluation for clinical applications. Our methodology, grounded in radiologist feedback and validated question generation, offers a scalable framework for creating trustworthy benchmarks in other medical domains, laying the groundwork for evaluating reasoning over classification in future healthcare AI systems.

\paragraph{Limitation.}
While ReXVQA demonstrates strong performance as a benchmark, several limitations should be acknowledged. Although we addressed demographic diversity by integrating data from four different U.S. hospital systems, the dataset may not fully represent global radiological practices or diverse international patient populations. Furthermore, our evaluation focused primarily on eight models, with limited commercial representation (only Gemini 1.5 Pro included), leaving assessment of other commercial multimodal systems as an important area for future work. While ReXVQA does employ predetermined answer choices, it differs from existing classification systems by evaluating diverse cognitive reasoning patterns rather than simple disease label prediction. Future work should explore incorporating open-ended question formats to further assess flexible clinical reasoning. These limitations present opportunities for expanding benchmark coverage across more diverse demographic settings and model architectures.



\section*{Acknowledgement}
This work was supported by the Biswas Family Foundation’s Transformative Computational Biology Grant in Collaboration with the Milken Institute.





\bibliography{sample}

\newpage 
\appendix

\section{Appendix} 
\label{sec:appendix}

\begin{itemize}
  \item \textbf{LLaVA 1.5} (2023): An open-source chatbot model trained by fine-tuning LLaMA/Vicuna on GPT-generated multimodal instruction-following data. The model connects a CLIP ViT-L/14 visual encoder with Vicuna using a projection matrix and was trained on 158K unique language-image instruction samples \citep{llava2024}.
  
  \item \textbf{Phi-3.5-vision-instruct} (Microsoft, 2024): A lightweight multimodal model (4.2B parameters) supporting 128K token context length. Trained on 500B tokens across vision and text data, it excels at multi-frame understanding and image comparison tasks \citep{phi2vision2024}.
  
  \item \textbf{Qwen2-VL} (Alibaba, 2024): An open-source vision-language model with dynamic resolution capabilities that maintains original aspect ratios without distortion. \citep{qwenvl2024}.
  
  \item \textbf{Qwen2.5-VL} (Alibaba, 2025): An upgraded model with enhanced recognition of handwritten text and multiple languages. It introduces structured output capabilities for data extraction and accurate object localization through bounding boxes or points \citep{bai2025qwen2}.
  
  \item \textbf{Gemini 1.5 Pro} (Google, 2024): A multimodal model designed for complex reasoning with extensive context processing capabilities. It achieves near-perfect recall on long-context retrieval tasks and demonstrates significant improvements in document and video question answering \citep{geminiprov2024}.
  
  \item \textbf{Eagle2-9B} (NVIDIA, 2024): A vision-language model balancing performance and inference speed. It combines SigLip and ConvNext vision encoders with Qwen2.5-7B-Instruct \citep{li2025eagle}.
  
  \item \textbf{Janus-Pro-7B} (DeepSeek, 2025): An autoregressive framework unifying multimodal understanding and generation. Built on DeepSeek-LLM-7b-base with SigLIP-L vision encoder, it supports 384×384 image inputs and significantly outperforms its predecessor on multimodal understanding benchmarks \citep{chen2025janus}.

  \item \textbf{MedGemma}  (Google, 2025): MedGemma 4B utilizes a SigLIP image encoder that has been specifically pre-trained on a variety of de-identified medical data, including chest X-rays, dermatology images, ophthalmology images, and histopathology slides. Its LLM component is trained on a diverse set of medical data, including radiology images, histopathology patches, ophthalmology images, and dermatology images~\cite{medgemma-hf}.
  
\end{itemize}

\begin{table}[!ht]
\small
\centering
\begin{tabular}{lrr}
\toprule
{\bf Criteria} & {\bf Long-Form} & {\bf MCQ} \\
\midrule
Reproducibility & Limited \textcolor{red}{$-$} & High \textcolor{green}{$+$} \\
Standardization & Variable \textcolor{red}{$-$} & Consistent \textcolor{green}{$+$} \\
Quantification & Subjective \textcolor{red}{$-$} & Objective \textcolor{green}{$+$} \\
Resources & High \textcolor{red}{$-$} & Low \textcolor{green}{$+$} \\
Scalability & Limited \textcolor{red}{$-$} & Extensive \textcolor{green}{$+$} \\
Granularity & Coarse \textcolor{red}{$-$} & Fine \textcolor{green}{$+$} \\
Reasoning & Implicit \textcolor{red}{$-$} & Explicit \textcolor{green}{$+$} \\
Automation & Limited \textcolor{red}{$-$} & High \textcolor{green}{$+$} \\
\bottomrule
\end{tabular}
\caption{Comparison of evaluation methodologies for radiological LLM assessment. \textcolor{green}{$+$} indicates advantage, \textcolor{red}{$-$} indicates limitation.}
\label{tab:comparison12}
\vspace{-2ex}
\end{table}

\begin{figure*}[!ht]
    \centering
    \includegraphics[width=\textwidth, height=19cm, keepaspectratio]{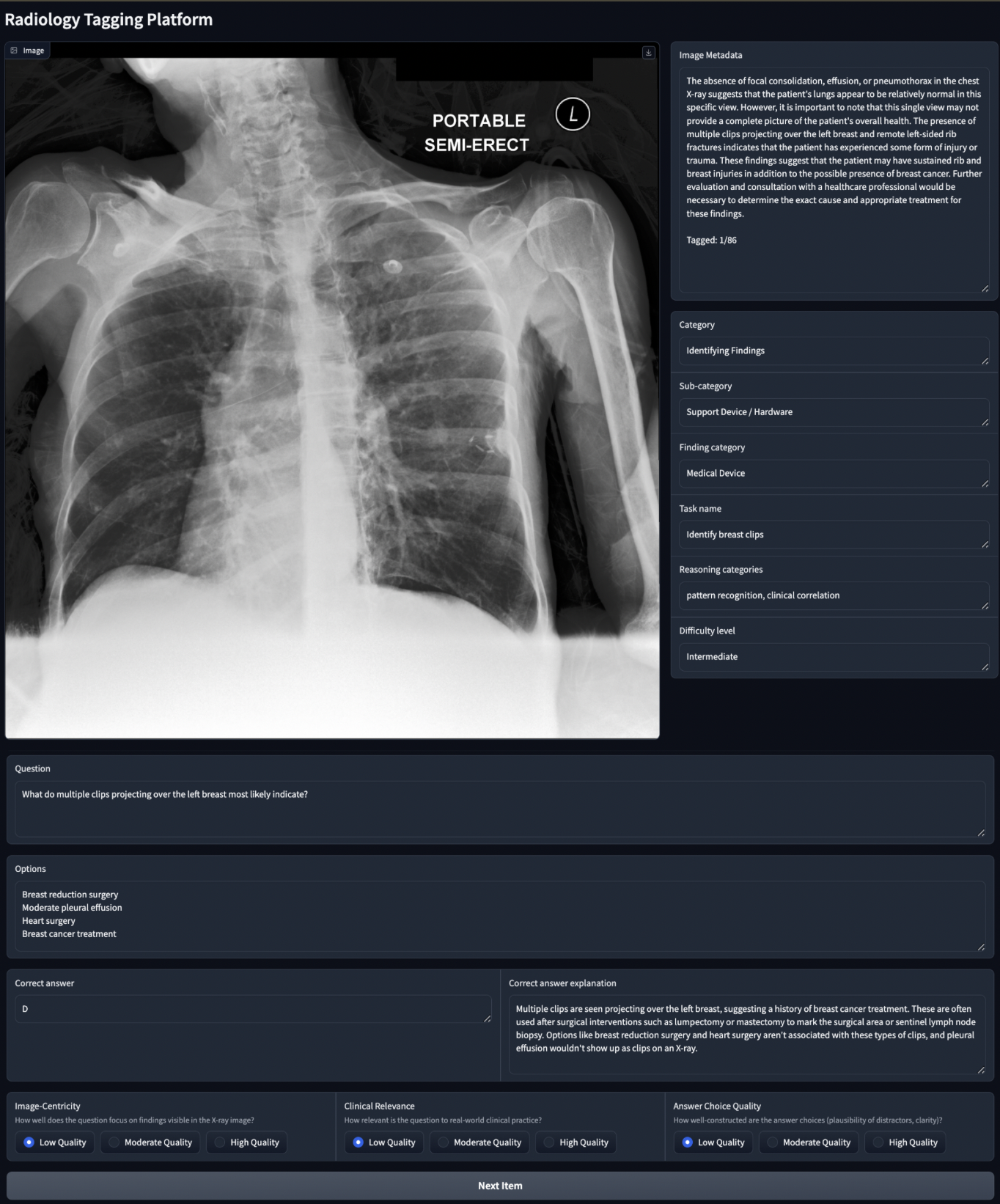}
    \caption{\textbf{The radiology image tagging platform interface used for expert annotation.} The platform displays a portable chest X-ray with associated metadata, categorization fields, and multiple-choice assessment options. The interface includes structured fields for capturing finding categories, difficulty levels, and clinical reasoning, alongside expert feedback on image-centricity and clinical relevance. This platform facilitated systematic collection of radiologist annotations and assessments for training data validation.}
    \label{fig:radiolologyplatform_figure}
\end{figure*}

\begin{table*}[t]
\centering
\small
\begin{tabular}{p{0.3\linewidth}p{0.64\linewidth}}
\toprule
\textbf{Main Category} & \textbf{Subcategories} \\
\midrule
Congenital Disease & Congenital Lung Disease, Congenital Vascular Disease, Congenital Heart Disease \\
\midrule
Infectious Disease & Pneumonia, Tuberculosis, Other Infection \\
\midrule
Pulmonary Neoplasm & Primary Lung Cancer, Pulmonary Metastases, Other Pulmonary Neoplasm \\
\midrule
Lymphoproliferative Disease & Lymphoma, Other Lymphoproliferative Disease \\
\midrule
Other Pulmonary Diagnosis & Interstitial Lung Disease, Sarcoidosis, Asbestos-Related Disease, Pneumoconiosis, Pulmonary Edema, ARDS, Aspiration, Iatrogenic Lung Disease, COPD, Vasculitis, Pulmonary Hypertension, Pulmonary Thromboembolic Disease, Miscellaneous Pulmonary Disease \\
\midrule
Cardiac Disease & Valvular Heart Disease, Myocardial Disease, Pericardial Disease, Congestive Heart Failure, Other Cardiac Disease \\
\midrule
Aortic Disease & Aortic Dissection/Aneurysm, Atherosclerosis, Other Aortic Disease \\
\midrule
Miscellaneous Diagnosis & Trauma, Post-Treatment Change, Miscellaneous Disease \\
\midrule
Negation & Absence of Disease \\
\bottomrule
\end{tabular}
\caption{Classification of Medical Conditions and their Subcategories}
\label{tab:disease-classification}
\end{table*}

\begin{table*}[t]
\tiny
\centering
\setlength{\tabcolsep}{1pt}
\begin{tabular}{l|rrrrrrrr}
\toprule
\textbf{Category} & \textbf{Gemini} & \textbf{Eagle2} & \textbf{Janus-Pro-7B} & \textbf{LLaVA} & \textbf{Qwen2VL} & \textbf{Qwen25VL} & \textbf{Phi35} & \textbf{MedGemma} \\
\midrule
Abdomen & 75.34 & 55.48 & 52.05 & 28.77 & 59.59 & 61.64 & 32.50 & 78.77\\
Airway & 69.84 & 71.28 & 56.66 & 24.54 & 60.84 & 59.01 & 81.61 & 92.17 \\
Aorta & 76.65 & 41.04 & 60.14 & 6.84 & 72.17 & 34.79 & 33.05 & 87.86 \\
Aortic Disease & 45.00 & 28.57 & 19.05 & 9.52 & 14.29 & 19.05 & 25.00 & 33.33\\
Bone & 87.99 & 89.82 & 86.67 & 31.93 & 91.93 & 88.77 & 80.41 & 94.04 \\
Bone Density & 50.00 & 50.00 & 0.00 & 100.00 & 100.00 & 0.00 & 0.00 & 50.00\\
Bones & 100.00 & 100.00 & 100.00 & 25.00 & 100.00 & 91.67 & 87.50 & 100.00 \\
Bones and Soft Tissues & 100.00 & 100.00 & 100.00 & 100.00 & 100.00 & 100.00 & - & 100.00 \\
Bones/Joints & 100.00 & 100.00 & 100.00 & 100.00 & 100.00 & 100.00 & 100.00 & 100.00 \\
Bony Structures & 100.00 & 100.00 & 100.00 & 60.00 & 100.00 & 100.00 & 100.00 & 100.00 \\
Bony Thorax & 100.00 & 100.00 & 100.00 & 100.00 & 100.00 & 100.00 & 100.00 & 100.00 \\
Cardiac Disease & 59.03 & 80.69 & 28.97 & 30.56 & 33.10 & 95.17 & 23.62 & 80.56\\
Chest Wall & 60.00 & 30.00 & 70.00 & 30.00 & 60.00 & 70.00 & 37.50 & 40.00\\
Clavicle & 75.93 & 57.14 & 75.00 & 23.21 & 71.43 & 33.93 & 51.02 & 92.73\\
Congenital Disease & 71.43 & 71.43 & 57.14 & 40.00 & 71.43 & 42.86 & 33.33 & 57.14\\
Diaphragm & 64.16 & 51.18 & 40.83 & 8.61 & 61.83 & 42.60 & 44.24 & 68.93 \\
Gastrointestinal Devices & 16.67 & 33.33 & 66.67 & 50.00 & 33.33 & 16.67 & 25.00 & 60.00\\
Heart & 80.29 & 81.71 & 72.01 & 26.01 & 60.31 & 84.83 & 62.73 & 97.03\\
Hila & 100.00 & 100.00 & 0.00 & 0.00 & 0.00 & 100.00 & 100.00 & 100.00 \\
Hilar Opacity & 100.00 & 100.00 & 100.00 & 0.00 & 100.00 & 100.00 & 100.00 & 100.00 \\
Humerus & 67.35 & 56.86 & 82.35 & 39.22 & 66.67 & 47.06 & 52.38 & 84.00 \\
Implanted Devices & 54.46 & 52.64 & 60.79 & 29.50 & 52.40 & 53.48 & 43.16 & 73.14\\
Infectious Disease & 71.55 & 63.24 & 66.77 & 56.63 & 51.02 & 67.74 & 42.13 & 77.61\\
Joint & 51.96 & 42.31 & 70.19 & 47.06 & 66.35 & 36.54 & 42.22 & 88.46\\
Lung Volume & 65.44 & 58.22 & 51.64 & 39.72 & 52.11 & 60.92 & 28.25 & 78.64\\
Lung and Pleural Lucency & 80.00 & 64.65 & 58.59 & 19.34 & 61.62 & 78.64 & 58.82 & 87.88\\
Lung and Pleural Opacity & 72.24 & 72.19 & 68.70 & 32.98 & 60.73 & 64.77 & 58.14 & 80.44\\
Lymph Nodes & 100.00 & 11.11 & 0.00 & 0.00 & 0.00 & 100.00 & 0.00 & 22.22\\
Lymphoproliferative Disease & 90.00 & 60.00 & 50.00 & 30.00 & 60.00 & 80.00 & 57.14 & 60.00\\
Mediastinum & 93.27 & 79.77 & 85.93 & 15.24 & 70.85 & 90.83 & 80.73 & 90.58 \\
Miscellaneous Bone Abnormality & 52.43 & 50.49 & 62.62 & 39.32 & 65.85 & 43.20 & 40.74 & 74.15\\
Miscellaneous Diagnosis & 68.83 & 64.94 & 72.73 & 31.51 & 61.04 & 68.83 & 55.17 & 79.22\\
Musculoskeletal & 0.00 & 0.00 & 100.00 & 100.00 & 0.00 & 0.00 & 0.00 &100.00 \\
Neck/Chest Wall Soft Tissue & 85.94 & 77.27 & 56.06 & 27.27 & 66.67 & 51.52 & 41.38 & 83.33\\
Negation & 61.05 & 71.73 & 58.19 & 15.00 & 56.37 & 61.49 & 78.96 & 74.76\\
Non-Therapeutic Internal Foreign Bodies & 67.86 & 63.33 & 65.00 & 26.67 & 65.00 & 66.67 & 53.85 & 74.58\\
Osseous Structures & 95.24 & 100.00 & 100.00 & 23.81 & 100.00 & 100.00 & 100.00 & 100.00 \\
Osseous structures and Soft Tissues & 100.00 & 100.00 & 100.00 & 0.00 & 100.00 & 100.00 & 100.00 & 100.00 \\
Other (Artifacts and External Foreign Bodies) & 50.00 & 50.00 & 0.00 & 0.00 & 0.00 & 50.00 & 0.00 & 100.00 \\
Other Bone Abnormality & 0.00 & 0.00 & 0.00 & 100.00 & 0.00 & 100.00 & 0.00 & 100.00 \\
Other Great Vessel & 73.33 & 60.00 & 60.00 & 13.33 & 60.00 & 60.00 & 45.45 & 73.33 \\
Other Implanted Devices & 100.00 & 100.00 & 0.00 & 0.00 & 0.00 & 100.00 & 0.00 & 100.00 \\
Other Pulmonary Diagnosis & 59.85 & 59.85 & 45.51 & 22.07 & 43.00 & 64.77 & 31.14 & 81.26 \\
Pleura & 100.00 & 100.00 & 100.00 & 0.00 & 33.33 & 100.00 & 100.00 & 100.00 \\
Pleural & 100.00 & 0.00 & 0.00 & 0.00 & 100.00 & 0.00 & 0.00 & 0.00 \\
Pleural Effusion & 87.88 & 93.94 & 42.42 & 6.06 & 39.39 & 84.85 & 75.00 & 69.70 \\
Pleural Space & 100.00 & 100.00 & 100.00 & 0.00 & 100.00 & 100.00 & 0.00 & 100.00 \\
Pleural Thickening & 79.84 & 75.78 & 67.97 & 23.44 & 58.59 & 71.88 & 57.43 & 82.81 \\
Pulmonary Fissure & 90.00 & 60.00 & 80.00 & 40.00 & 60.00 & 30.00 & 75.00 & 90.00 \\
Pulmonary Neoplasm & 78.46 & 66.92 & 81.95 & 39.29 & 63.91 & 73.68 & 69.44 & 88.72 \\
Pulmonary Vascularity & 73.30 & 67.88 & 82.39 & 25.54 & 38.31 & 78.63 & 50.44 & 83.31 \\
Rib & 88.90 & 83.93 & 89.80 & 36.92 & 86.35 & 79.21 & 78.56 & 91.84 \\
Scapula & 28.57 & 14.29 & 42.86 & 28.57 & 28.57 & 14.29 & 33.33 & 57.14 \\
Shoulder & 75.00 & 50.00 & 50.00 & 50.00 & 100.00 & 50.00 & 0.00 & 100.00 \\
Skeletal Structures & 98.86 & 99.62 & 99.87 & 32.83 & 100.00 & 99.87 & 99.82 & 100.00 \\
Skeletal System & 70.00 & 90.00 & 80.00 & 80.00 & 90.00 & 70.00 & 87.50 & 100.00 \\
Skeleton & 100.00 & 100.00 & 100.00 & 75.00 & 100.00 & 100.00 & 100.00 & 100.00 \\
Soft Tissue & 60.00 & 80.00 & 80.00 & 20.00 & 60.00 & 80.00 & 25.00 & 100.00 \\
Spine & 78.84 & 62.30 & 86.43 & 64.84 & 73.73 & 50.10 & 57.94 & 92.68\\
Sternum & 92.50 & 85.37 & 87.80 & 70.00 & 87.80 & 87.80 & 75.00 & 92.68\\
Trauma & 60.00 & 60.00 & 100.00 & 60.00 & 60.00 & 80.00 & 80.00 & 60.00 \\
Tubes and Lines & 59.45 & 58.26 & 58.87 & 22.81 & 48.78 & 65.04 & 32.10 & 83.86 \\
Vascular & 0.00 & 100.00 & 100.00 & 100.00 & 100.00 & 0.00 & 0.00 & 100.00 \\
Vasculature & 100.00 & 100.00 & 100.00 & 100.00 & 100.00 & 100.00 & 0.00 & 100.00\\
Vascularity & - & 0.00 & 100.00 & 100.00 & 100.00 & 0.00 & 0.00 & 100.00\\
\midrule
\textbf{Average} & 74.00 & 67.00 & 67.00 & 38.00 & 64.00 & 66.00 & 50.00 & 83.24 \\
\bottomrule
\end{tabular}
\caption{Performance comparison of various models across different categories (all values in \%). The average is calculated across all categories. }
\label{tab:model_comparison}
\end{table*}

\section{Data Availability}

The ReXVQA benchmark will be made publicly available to the research community. This includes:

\begin{itemize}
    \item \textbf{Dataset}: The public portions of ReXVQA containing training (572,952), validation (40,878), and public test (40,826) question-answer pairs with 160,000 chest X-rays. The private test set remains confidential for unbiased leaderboard evaluation.


    \item \textbf{Leaderboard}: A public leaderboard system where researchers can submit predictions on the private test set for independent evaluation
\end{itemize}
The benchmark will be hosted at \href{https://rexrank.ai/}{rexrank.ai} for research purposes.

\end{document}